\documentclass[sigconf]{acmart}
\AtBeginDocument{%
  }

\copyrightyear{2027}
\acmYear{2027}
\setcopyright{cc}
\setcctype{by}
\acmConference[WSDM '27]{Proceedings of the Twentieth ACM International Conference on Web Search and Data Mining}{February 15--19, 2027}{Hong Kong, Hong Kong}
\acmBooktitle{Proceedings of the Twentieth ACM International Conference on Web Search and Data Mining (WSDM '27), February 15--19, 2027, Hong Kong, Hong Kong}
\acmDOI{10.1145/3862322.3863579}
\acmISBN{979-8-4007-3117-4/2027/02}

\usepackage{multirow}
\usepackage{hyperref}

\begin{document}

\title{RelPrism: Multi-Faceted Self-Supervised Learning on Relational Databases}

\author{Jinyu Yang}
\orcid{0009-0007-5467-0690}
\affiliation{%
  \institution{Beijing University of Posts and Telecommunications}
  \city{Beijing}
  \country{China}
}
\affiliation{%
  \institution{Peng Cheng Laboratory}
  \city{Shenzhen}
  \country{China}
}
\email{jinyu.yang@bupt.edu.cn}

\author{Cheng Yang}
\orcid{0000-0001-7821-0030}
\affiliation{%
  \institution{Beijing University of Posts and Telecommunications}
  \city{Beijing}
  \country{China}
}
\email{yangcheng@bupt.edu.cn}

\author{Junze Chen}
\orcid{0009-0001-9619-5492}
\affiliation{%
  \institution{China Telecom Bestpay}
  \city{Beijing}
  \country{China}
}
\email{chenjunze@besypay.com.cn}

\author{Zedi Liu}
\orcid{0009-0003-2022-3594}
\affiliation{%
  \institution{Beijing University of Posts and Telecommunications}
  \city{Beijing}
  \country{China}
}
\email{liuzedi@bupt.edu.cn}


\author{Muhan Zhang}
\orcid{0000-0002-7680-6401}
\affiliation{%
  \institution{Peking University}
  \city{Beijing}
  \country{China}
}
\email{muhan@pku.edu.cn}

\author{Hanyang Peng}
\authornote{Corresponding authors.}
\orcid{0000-0002-9715-473X}
\affiliation{%
  \institution{Peng Cheng Laboratory}
  \city{Shenzhen}
  \country{China}
}
\email{penghy@pcl.ac.cn}

\author{Chuan Shi}
\authornotemark[2]
\orcid{0000-0002-3734-0266}
\affiliation{%
  \institution{Peng Cheng Laboratory}
  \city{Shenzhen}
  \country{China}
}
\affiliation{%
  \institution{Beijing University of Posts and Telecommunications}
  \city{Beijing}
  \country{China}
}
\email{shichuan@bupt.edu.cn}

\renewcommand{\shortauthors}{Jinyu Yang et al.}

\begin{abstract}
  Relational databases (RDBs) remain the backbone of modern data systems and support a broad spectrum of predictive tasks that require multi-faceted information from different perspectives and granularities.
  For example, user churn prediction is often formulated as a binary classification task that relies more on user interaction patterns reflecting activity levels, whereas consumption value estimation is a fine-grained regression task that may require composite evidence from both user attributes and user--item interaction behaviors.
  Therefore, a central challenge is to learn representations with comprehensive coverage of the information underlying RDB tasks.
  A natural solution is to leverage self-supervised learning (SSL) for pre-training.
  However, existing SSL methods typically derive supervision from a single facet, such as node-level intrinsic attributes or subgraph-level relational structures.
  Consequently, their pre-training objectives provide limited support for adaptation.
  To this end, we propose RelPrism, a multi-faceted self-supervised learning framework for RDBs.
  Concretely, RelPrism first constructs intrinsic, relational, and hybrid attributes from distinct perspectives. It then applies multi-granularity clustering to each attribute perspective to form corresponding pseudo-task pools.
  Pre-training over these task pools exposes representations to comprehensive attribute perspectives and granularity levels, providing a stronger basis for downstream adaptation.
  Extensive experiments on 14 tasks across 5 real-world datasets show that RelPrism improves ROC-AUC by 4.15\% on classification tasks and
  reduces MAE by 10.75\% on regression tasks over state-of-the-art baselines.
  Our code is available at \url{https://github.com/BUPT-GAMMA/RelPrism}.
\end{abstract}

\ccsdesc[500]{Computing methodologies~Unsupervised learning}
\ccsdesc[500]{Information systems~Data mining}

\keywords{Relational Databases; Self-Supervised Learning}


\maketitle

\section{Introduction}
\begin{figure}[t]
    \centering
    \includegraphics[width=1\linewidth]{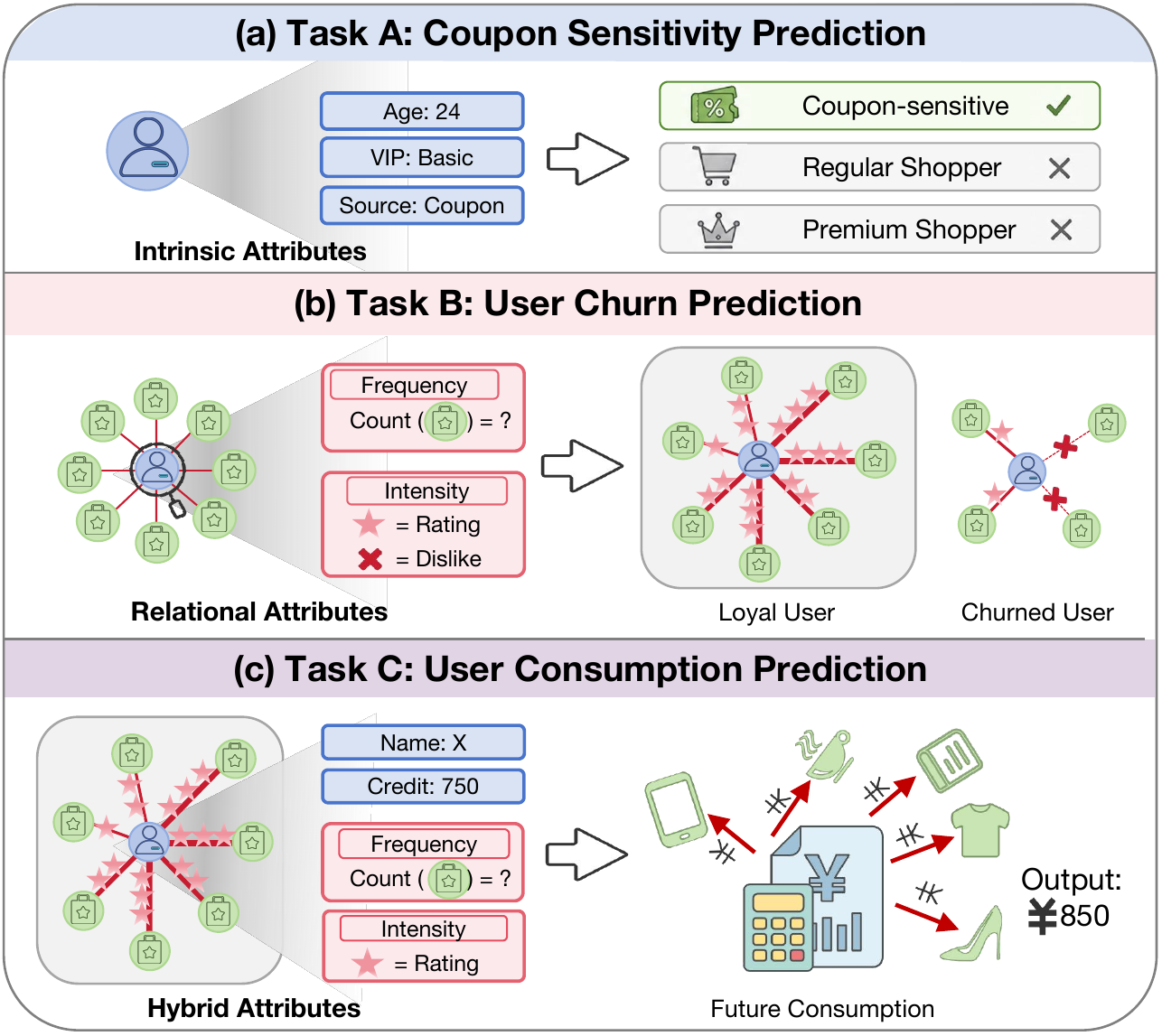}
    \caption{\textbf{RDB Predictive Tasks Require Multi-Faceted Information Across Perspectives and Granularities.} We illustrate three downstream tasks in a real-world e-commerce RDB. Specifically, (a) relies primarily on node-level intrinsic attributes, such as user source; (b) depends more on subgraph-level relational attributes that capture interaction patterns, such as interaction frequency and intensity; and (c) requires hybrid attributes to support fine-grained regression.}
    \label{fig:intro}
\end{figure}

Relational Databases (RDBs) serve as the operational substrate of many real-world systems, organizing high-value structured data across domains such as finance~\cite{dong2025transaction} and healthcare~\cite{ulmer2020trust}.
An RDB typically consists of multiple interconnected tables, where each table contains heterogeneous columns, including numerical, categorical, textual, and temporal attributes.
These structured relational databases often support a broad range of predictive tasks~\cite{robinson2024relbench,wang20244dbinfer}, such as user churn prediction and user consumption value prediction, as illustrated in Figure~\ref{fig:intro}.
Conventional predictive modeling methods for RDBs often rely on manual feature engineering to flatten multi-table information into feature vectors~\cite{banachewicz2022kaggle}, which is labor-intensive, requires domain expertise, and may introduce biases that limit performance~\cite{bengio2013representation}. 
In recent years, Relational Deep Learning (RDL)~\cite{fey2023relational} has emerged as a powerful end-to-end learning paradigm tailored for RDBs.
By converting RDBs into temporal heterogeneous graphs, where rows serve as nodes and inter-table relations (such as primary-foreign keys) serve as edges, RDL provides a graph-based formulation for relational predictive tasks.

Despite the promise of RDL for end-to-end learning, designing effective self-supervised learning (SSL) methods for RDBs remains non-trivial~\cite{truong2025pre}. 
Predictive tasks over RDBs often require multi-faceted information, involving different attribute perspectives and prediction granularities. 
As illustrated in Figure~\ref{fig:intro}, in a real-world e-commerce RDB, Task A may primarily rely on intrinsic attributes such as user source, Task B may depend more on relational attributes, where dense and positive interactions indicate user loyalty, while Task C may require hybrid evidence to estimate continuous consumption value.
Therefore, self-supervised pre-training for RDBs should expose representations to comprehensive information across perspectives and granularities, enabling the learned representations to better support adaptation.

However, existing SSL methods typically utilize single-faceted information as supervision during pre-training.
When applied to RDBs, single-table SSL methods~\cite{yoon2020vime,nam2023stunt,bahri2021scarf} mainly focus on row-level intrinsic attributes, for example by corrupting part of a row and reconstructing its feature values~\cite{yoon2020vime}.
Such objectives capture only a single intrinsic perspective and overlook the rich relational dependencies in RDBs.
Another line of work follows the RDL paradigm by converting RDBs into graphs and employing graph SSL objectives~\cite{you2020graph,hou2022graphmae}, but directly applying them can lead to unstable or even negative transfer~\cite{liu2023flaky}, as the learned signals may not consistently align with downstream relational prediction.
A recent RDB-oriented pre-training method~\cite{truong2025pre} explicitly considers relational structures, yet its supervision is still mainly derived from subgraph-level relational attributes and remains tied to a particular perspective and granularity.
Since downstream information needs are unknown during pre-training, such single-faceted objectives may provide insufficient coverage of the information space, limiting the adaptability of learned representations.

To address these challenges, we propose RelPrism, a multi-faceted self-supervised learning framework tailored for RDBs.
The core idea is to generate pseudo-tasks that expose representations to broader attribute perspectives and multiple granularity levels for pre-training, thereby building a more comprehensive foundation for downstream adaptation.
Specifically, we first construct attributes from multiple perspectives, including intrinsic attributes derived from each node's raw row, relational attributes that characterize diverse interaction patterns around each node, and their hybrid fusion that captures composite signals beyond any single perspective.
These complementary attributes provide comprehensive node profiling and serve as the basis for subsequent pseudo-task generation.
Then, we adopt a multi-granularity clustering strategy on each perspective's attributes to generate pseudo-tasks with different prediction granularities and construct their corresponding task pools.
Pre-training over these task pools encourages RelPrism to encode multi-faceted information, yielding representations that are better prepared for downstream adaptation.

Our main contributions are summarized as follows:

$\bullet$ We identify a key mismatch between existing single-faceted self-supervised objectives and the comprehensive information coverage required for RDB pre-training, showing that effective representation learning should account for both multiple attribute perspectives and multiple predictive granularities.

$\bullet$ We propose RelPrism, a multi-faceted SSL framework for RDBs that constructs intrinsic, relational, and hybrid attributes from complementary perspectives, and generates corresponding pseudo-task pools through multi-granularity clustering for pre-training.

$\bullet$ We conduct extensive experiments on 14 tasks across 5 real-world datasets spanning both data-limited and data-sufficient scenarios, demonstrating that RelPrism outperforms state-of-the-art baselines by an average of 4.15\% in ROC-AUC for classification and 10.75\% in MAE for regression.

\section{Related Work}
\subsection{Self-Supervised Learning on Tabular Data}
Tabular data constitutes the fundamental unit of RDBs, corresponding to individual tables within the RDBs~\cite{chen2016xgboost,borisov2022deep}.
Tabular SSL methods leverage extensive unlabeled data to design pretext tasks~\cite{nam2023stunt} for pre-training, such as denoising reconstruction~\cite{yoon2020vime,ucar2021subtab} or multi-view contrastive learning~\cite{bahri2021scarf,somepalli2021saint}. For instance, VIME~\cite{yoon2020vime} estimates mask vectors from corrupted tabular inputs for reconstruction, while SCARF~\cite{bahri2021scarf} constructs diverse views for contrastive learning by corrupting random feature subsets. However, despite their success in single-table scenarios, these methods focus primarily on node-level intrinsic attributes and overlook the relational dependencies inherent in RDBs, which critically limits their effectiveness for downstream predictive tasks~\cite{kanter2015deep,schlichtkrull2018modeling}.

\subsection{Self-Supervised Learning on RDB Graphs}
Considering the rich relational dependencies within RDBs, Relational Deep Learning (RDL)~\cite{fey2023relational} has gained increasing attention in recent years.
These approaches typically convert RDBs into \textit{temporal heterogeneous graphs}, where rows are represented as nodes and inter-table relations (such as primary-foreign keys) are represented as edges.
Graph Neural Networks (GNNs) are then employed to learn representations over the resulting graphs~\cite{cvitkovic2020supervised,cvetkov2023relational,chen2025relgnn,dwivedi2025relational1,dwivedi2025relational2}.
Given this graph-based formulation, a straightforward SSL strategy is to directly adopt existing graph self-supervised learning objectives~\cite{you2020graph,hou2022graphmae,velivckovic2018deep} on RDB-induced graphs.
Nevertheless, prior studies have shown that such naive application can be counterproductive, often leading to the learning of \textit{flaky information} that does not consistently benefit downstream relational prediction~\cite{liu2023flaky}.

Although a recent RDB-oriented self-supervised learning method utilizes subgraph-level relational structures as supervision~\cite{truong2025pre} during pre-training, relying solely on topology offers limited single-faceted information, which is insufficient for adaptation. In addition, recent RDB foundation models~\cite{wang2025griffin,fey2025kumorfm,ranjan2025relational} which leverage pre-training on large-scale corpora to generalize across a wide range of tasks, are orthogonal and complementary to the pre-training framework and strategy investigated in this work. Nevertheless, we provide detailed empirical comparisons in Section~\ref{sec:experiments and analysis}.

\section{Preliminaries}
\label{sec:preliminary}
\begin{figure*}[t]
    \centering
    \includegraphics[width=1\linewidth]{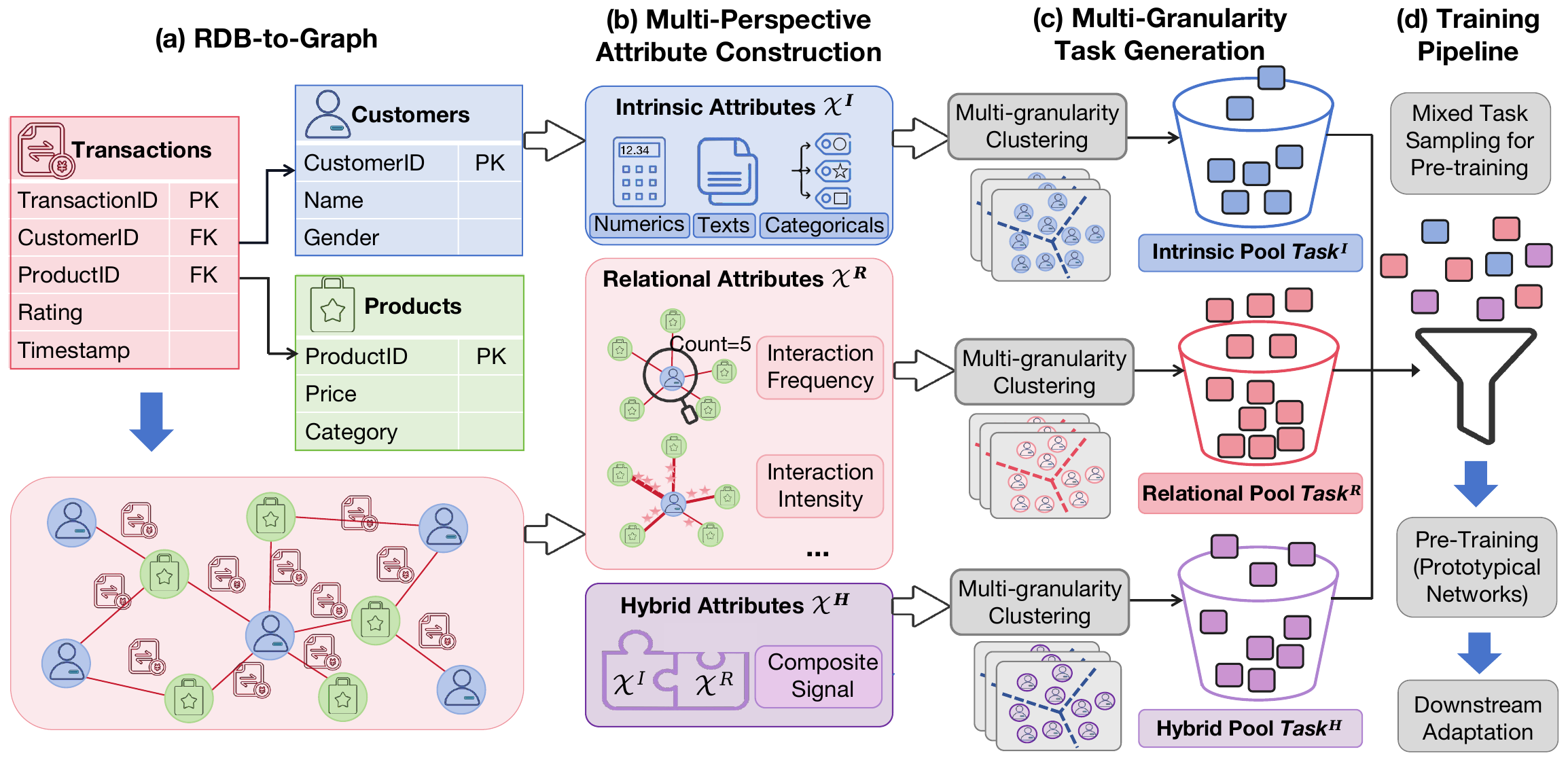}
    \caption{\textbf{The Overall Framework of RelPrism.}
    (a) We first convert the RDB into a temporal heterogeneous graph. 
    (b) Next, we construct multi-perspective node attributes: intrinsic attributes $\mathcal{X}^I$ derived from the corresponding row's raw data, relational attributes $\mathcal{X}^R$ extracted from interaction information, and hybrid attributes that fuse both to encode composite information.
    (c) We then apply a multi-granularity clustering strategy independently to each perspective to construct corresponding multi-granularity task pools: the Intrinsic Pool $Task^I$, the Relational Pool $Task^R$, and the Hybrid Pool $Task^H$.
    (d) Finally, we perform mixed task sampling from these task pools for pre-training and then obtain representations for downstream adaptation.
   }
    \label{fig:framework}
\end{figure*}

\textit{\textbf{Relational Databases.}}
A relational database consists of multiple interconnected tables. Formally, we denote it as $\mathcal{D} = \mathcal{T}_{\text{dim}} \cup \mathcal{T}_{\text{fact}}$, where $\mathcal{T}_{\text{dim}}$ and $\mathcal{T}_{\text{fact}}$ are the set of dimension tables and fact tables, respectively.
Dimension tables are typically static, storing descriptive attributes (such as Customer, Product in Figure~\ref{fig:framework}(a)), and are referenced by foreign keys in fact tables or other dimension tables. In contrast, fact tables record dynamic interactions between dimension tables (such as Transactions in Figure~\ref{fig:framework}(a)), contain multiple foreign key pairs that reference dimension tables, and typically have no incoming references. 
Each table $T \in \mathcal{D}$ contains multiple rows and columns.
The columns are often heterogeneous, including numerical, textual, categorical, and temporal types. Some columns serve as foreign keys, establishing relational dependencies: (i) When a foreign key (FK) in dimension table $T \in \mathcal{T}_{\text{dim}}$ references the primary key (PK) of another dimension table $T' \in \mathcal{T}_{\text{dim}}$, a primary-foreign key (PK-FK) dependency is established between them. (ii) When a pair of foreign keys in table $T \in \mathcal{T}_{\text{fact}}$ references the PK of dimension tables $T' \in \mathcal{T}_{\text{dim}}$ and $T'' \in \mathcal{T}_{\text{dim}}$, a foreign-foreign key (FK-FK) dependency is established between them.

\textit{\textbf{Problem Definition.}}
Following prior work~\cite{fey2023relational,robinson2024relbench,chen2025relgnn}, we focus on common node-level predictive tasks over RDBs, including classification and regression. 
For each row in the target table $T_m \in \mathcal{T}_{\mathrm{dim}}$, which is treated as a target node, our goal is to encode its raw attributes and corresponding relational dependencies into a $d$-dimensional latent representation. 
These representations are then used for downstream adaptation on predictive tasks defined over the target table.
For downstream adaptation, we consider two regimes: (i) \textit{data-limited scenarios} and (ii) \textit{data-sufficient scenarios}. 
In contrast to prior work~\cite{truong2025pre}, our data-limited setting does not restrict the target population. 
Instead, it follows a standard few-shot protocol~\cite{hegselmann2023tabllm}, where only the number of labeled examples is restricted.
Specifically, it is formalized as $k$-shot adaptation, where $k$ denotes the number of labeled examples per class for classification tasks, or the total number of labeled examples for regression tasks. Data-sufficient scenarios correspond to the original RelBench~\cite{robinson2024relbench} setting, where sufficient supervision is available for adaptation.


\section{Methodology}
\subsection{Overview}
This section presents RelPrism, with its framework illustrated in Figure~\ref{fig:framework}.
By generating pseudo-tasks that span complementary attribute perspectives and multiple granularity levels for pre-training, RelPrism encourages the encoder to capture broader and more comprehensive information for downstream adaptation.
Specifically, RelPrism first converts the RDB into a temporal heterogeneous graph (Figure~\ref{fig:framework}(a)), where each row in a dimension table is represented as a node instance, and inter-table interactions are represented as edge instances.
Next, it constructs multi-perspective node attributes (Figure~\ref{fig:framework}(b)), including intrinsic attributes $\mathcal{X}^I$, relational attributes $\mathcal{X}^R$, and hybrid attributes $\mathcal{X}^H$.
By applying clustering multiple times with varying cluster counts to each perspective, RelPrism generates multi-granularity task pools (Figure~\ref{fig:framework}(c)): the Intrinsic Pool $Task^{I}$, Relational Pool $Task^{R}$, and Hybrid Pool $Task^{H}$.
This automated process resembles the label construction logic commonly used in data science, such as segmenting users based on consumption behaviors.
Finally, we employ Prototypical Networks~\cite{snell2017prototypical} for pre-training with a mixed task sampling strategy, enabling effective adaptation to target downstream tasks (Figure~\ref{fig:framework}(d)).

\subsection{RDB-to-Graph}
\label{sec:rdb2graph}
We formalize the relational database $\mathcal{D}$ as a \emph{temporal heterogeneous graph} $\mathcal{G} = (\mathcal{V}, \mathcal{E}, \mathcal{A}, \mathcal{R})$, where $\mathcal{A}$ and $\mathcal{R}$ denote the node and edge type sets~\cite{wang2019heterogeneous}.
The common RDL graph construction~\cite{fey2023relational} treats both dimension-table rows and fact-table rows as nodes, which can introduce redundant aggregations through intermediate fact nodes~\cite{chen2025relgnn}. 
To mitigate this issue, following prior work~\cite{wang20244dbinfer}, we model only rows in dimension tables as nodes, while representing inter-table interactions (the FK-FK and PK-FK dependencies) as edges, as illustrated in Figure~\ref{fig:framework}(a).

$\bullet$ \textbf{Node Sets.} We model each row in a dimension table $T \in \mathcal{T}_{\text{dim}}$ as a node $v$, denoting this row as the \textit{corresponding row} $row^{\mathcal{V}}_v$. Given that $row^{\mathcal{V}}_v$ consists of multiple columns, we let $row^{\mathcal{V}}_{v,j}$ denote the raw value of the $j$-th column.
All nodes derived from $T$ belong to the same node type $a \in \mathcal{A}$.
Therefore, the overall node set is given by $\mathcal{V} = \bigcup_{a \in \mathcal{A}} \mathcal{V}_a$.

$\bullet$ \textbf{Edge Sets.} Edges are derived from two distinct sources. First, given that FK-FKs in fact tables capture interactions between dimension tables, we directly model them as edges. Second, we represent the PK-FKs between dimension tables as edges. Formally:

(i) \textit{FK-FK Edges:} Each FK-FK pair in a fact table $T \in \mathcal{T}_{\text{fact}}$ is defined as an edge type $r \in \mathcal{R}$. Accordingly, every row in $T$ containing this pair constitutes an edge $e \in \mathcal{E}_r$ connecting the referenced nodes in two dimension tables. Here, the row itself is denoted as $row^{\mathcal{E}}_e$, and $row^{\mathcal{E}}_{e,j}$ represents the raw value of its $j$-th attribute.

(ii) \textit{PK-FK Edges:} Each PK-FK dependency between dimension tables is defined as a unique edge type $r \in \mathcal{R}$. Accordingly, an edge $e \in \mathcal{E}_r$ is derived from the primary key node to the foreign key node.
Since PK-FKs lack explicit entity rows as FK-FKs in fact tables, the \textit{corresponding row} $row^{\mathcal{E}}_e$ is defined as the row of the primary key node.
Collectively, the overall edge set is given by $\mathcal{E} = \bigcup_{r \in \mathcal{R}} \mathcal{E}_r$.

Additionally, all graph elements are timestamped, with the timestamp for each node $v \in \mathcal{V}$ and edge $e \in \mathcal{E}$ aligning with the creation time of its \textit{corresponding row}.

\subsection{Multi-Perspective Attribute Construction}
\label{sec:attribute_construction}

While prior data profiling studies focus on discovering (relaxed) functional dependencies in relational data~\cite{caruccio2015relaxed,breve2024decentralized,caruccio2023malicious}, we leverage the raw attributes and relational structures of $\mathcal{V}$ to construct node profiles comprising three complementary types of attributes: intrinsic attributes $\mathcal{X}^I$, relational attributes $\mathcal{X}^R$~\cite{neville2000iterative}, and hybrid attributes $\mathcal{X}^H$. These attributes provide multi-perspective information for subsequent task generation.

\paragraph{\textbf{Intrinsic Attributes.}}
We create intrinsic attributes for all nodes in $\mathcal{V}$ and all edges in $\mathcal{E}$ by encoding the raw data of their \textit{corresponding rows}.
For each node $v \in \mathcal{V}$, we consider a total number of $N_v$ raw input columns from $row_v^\mathcal{V}$ excluding PKs and FKs. Let $\phi_j(\cdot)$ denote the corresponding encoder for the $j$-th column, we obtain the intrinsic attribute $\mathbf{x}_{v}^I \in \mathbb{R}^{d_I}$ by concatenating the encoded results:
\begin{equation}
\mathbf{x}_{v}^I = \mathrm{Concat} \left( \phi_1(row_{v,1}^\mathcal{V}), \dots, \phi_{N_v}(row_{v,N_v}^\mathcal{V}) \right),
\end{equation}
where $\mathrm{Concat}$ denotes the concatenation operation, and consistent with~\cite{wang2025griffin}, $\phi_j$ is selected as a textual or numerical encoder based on the input type.

Similarly, for each edge $e \in \mathcal{E}$, we consider a total number of $N_e$ raw input columns from $row_{e}^\mathcal{E}$. Leveraging the same set of encoders, we obtain the intrinsic attribute $\mathbf{z}_{e}^I$:
\begin{equation}
\mathbf{z}_e^I = \mathrm{Concat} \left( \phi_1(row_{e,1}^\mathcal{E}), \dots, \phi_{N_e}(row_{e,N_e}^\mathcal{E}) \right)
\end{equation}

\paragraph{\textbf{Relational Attributes.}}
To capture the interaction patterns implicit in RDB relational dependencies, we construct a set of statistical descriptors for each node $v \in \mathcal{V}$. Specifically, rather than exhaustively modeling all possible relational statistics, we focus on three task-agnostic descriptors that capture complementary aspects of local relational behavior: how frequently a node interacts, how strong its interactions are, and with whom it interacts. Such statistics have been widely used as lightweight descriptors of local relational structure in graph representation learning~\cite{henderson2011s,henderson2012rolx}. 

Given a node $v \in \mathcal{V}$, let $\mathcal{U}_v$ denote the set of one-hop neighbors, where each neighbor $u \in \mathcal{U}_v$ is connected to $v$ via an edge $e=(u, v)$.
We derive the relational attributes $\mathbf{x}_v^R \in \mathbb{R}^{d_R}$ by aggregating statistics over $\mathcal{U}_v$ and corresponding edges across three aspects: interaction frequency $x_{v}^{R, freq}$, interaction intensity $x_{v}^{R, int}$, and neighborhood quality $x_{v}^{R, qual}$. For isolated nodes with $|\mathcal{U}_v|=0$, they are set to zero to indicate the absence of interactions.

$\bullet$ \textit{Interaction Frequency.} We utilize node degree to measure activity levels~\cite{faloutsos1999power}:
\begin{equation}
x_v^{R,freq} = \log(1 + |\mathcal{U}_v|)
\end{equation}

$\bullet$ \textit{Interaction Intensity.} 
To quantify this, we aggregate the intrinsic edge attributes for each edge $e=(u,v)$ connecting the target node $v$ to a neighbor $u \in \mathcal{U}_v$. 
We use the $L_2$ norm $\|\mathbf{z}_e^I\|_2$ as a compact scalar measure of edge intensity, computing both the log-sum and mean~\cite{corso2020principal} to capture the cumulative contribution and average behavior, respectively:
\begin{equation}
x_{v}^{R, int_{sum}} = \log\left(1 + \sum_{u \in \mathcal{U}_v} \|\mathbf{z}_{e=(u,v)}^I\|_2\right)
\end{equation}

\begin{equation}
x_{v}^{R,int_{mean}} = \frac{1}{|\mathcal{U}_v|} \sum_{u \in \mathcal{U}_v} \|\mathbf{z}_{e=(u,v)}^I\|_2
\end{equation}

Subsequently, these two components are concatenated to form the aggregate interaction intensity attribute $x_{v}^{R, \text{int}}$:
\begin{equation}
x_{v}^{R,int} = \text{Concat}(x_{v}^{R,int_{sum}}, x_{v}^{R,int_{mean}})
\end{equation}

$\bullet$ \textit{Neighborhood Quality.}
Beyond interaction frequency and intensity, the attributes of neighboring nodes provide important relational context for characterizing the target node~\cite{mcpherson2001birds}. For instance, a user frequently interacting with high-value items may exhibit stronger purchasing power. Therefore, we quantify neighborhood quality by aggregating the intrinsic attribute norms of all neighbor nodes. Formally, considering each neighbor $u \in \mathcal{U}_v$, we define:
\begin{equation}
x_{v}^{R,qual} = \log\left(1 + \text{MEAN}_{u \in \mathcal{U}_v} \left( \|\mathbf{x}_u^I\|_2 \right) \right)
\end{equation}

Finally, we concatenate these relational statistics to obtain the final relational attributes:
\begin{equation}
\mathbf{x}_{v}^R = \text{Concat}(x_{v}^{R,freq}, x_{v}^{R,int}, x_{v}^{R,qual})
\end{equation}

This process constructs lightweight, task-agnostic relational descriptors for nodes. These descriptors serve as a simple yet effective basis for subsequent pseudo-task generation and pre-training, while remaining interpretable and extensible.

\paragraph{\textbf{Hybrid Attributes.}}
To capture complementary signals beyond any single attribute perspective and better align with complex downstream task logic in RDBs, we fuse intrinsic and relational attributes to generate hybrid attributes $\mathbf{x}_v^H \in \mathbb{R}^{d_I+d_R}$. Formally:

\begin{equation}
\mathbf{x}_v^H = \text{Concat}(\mathbf{x}_v^I, \mathbf{x}_v^R)
\end{equation}

\subsection{Multi-Granularity Task Generation}
To cover the comprehensive informational needs required by a target predictive task, we generate pseudo-tasks from both multiple attribute perspectives and clustering granularities. This design follows two fundamental principles: ensuring a \emph{diverse and structured} task distribution while maintaining \emph{moderate difficulty}~\cite{hsu2018unsupervised}.

Specifically, focusing on the target node type $a \in \mathcal{A}$, we conduct $P$ independent clustering runs on the attributes of $\mathcal{V}_a$ under each perspective. 
For each run $p \in \{1, \dots, P\}$, we uniformly sample the number of clusters $C_p$ from a pre-defined range $[C_{\min}, C_{\max}]$, where $C_p$ controls the granularity of the generated pseudo-task.
Subsequently, we partition $\mathcal{V}_a$ into $C_p$ disjoint clusters based on the attributes of a specific perspective using K-Means~\cite{lloyd1982least}, following prior studies~\cite{hsu2018unsupervised,nam2023stunt} showing that K-Means-based pseudo-label construction provides a simple yet effective way to generate tasks from unlabeled data.
Finally, for each node $v \in \mathcal{V}_a$, we assign a pseudo-label based on its nearest centroid, which is denoted as $\tilde{y}_{v,p}^{I}$, $\tilde{y}_{v,p}^{R}$, or $\tilde{y}_{v,p}^{H}$ according to the current perspective. Formally, we define the pseudo-tasks $Task_{p}^{I}$, $Task_{p}^{R}$, and $Task_{p}^{H}$ generated in the $p$-th run as sets of input--label pairs.
For all pseudo-tasks, we use the hybrid node attributes as input to provide a unified and comprehensive node profile, while the pseudo-labels derived from different perspectives provide the corresponding supervision.
Taking $Task_{p}^{I}$ as an example, we have:
\begin{equation}
Task_{p}^{I} = \left\{ (\mathbf{x}_v^H, \tilde{y}_{v,p}^{I}) \mid v \in \mathcal{V}_a \right\}
\end{equation}

Collectively, we establish three task pools, each comprising $P$ pseudo-tasks derived from one attribute perspective. The full collection is denoted as $Task = \{Task^{I}, Task^{R}, Task^{H}\}$. These pools serve as a comprehensive task repository, providing a diverse and realistic task distribution for pre-training, enabling the model to learn representations that are adaptable to downstream tasks of target nodes with different perspective needs and granularity levels.

\subsection{Training Pipeline}
\label{training_pipeline}
In this subsection, we describe the training pipeline of RelPrism, which comprises pre-training and downstream adaptation. 
We pre-train a graph encoder $f_\theta$ on diverse self-generated tasks within a meta-learning~\cite{finn2017model} paradigm based on Prototypical Networks~\cite{snell2017prototypical}. 
Through episodic meta-training on multi-perspective and multi-granularity pseudo-tasks, $f_\theta$ learns representations that support rapid adaptation to downstream tasks.
Here, we implicitly incorporate the dimension alignment projectors for heterogeneous node and edge types into $f_\theta$ to simplify notation.

\paragraph{\textbf{Pre-Training.}}
First, we adopt a mixed sampling strategy to randomly sample tasks ${Task}_{pre}$ from task pools ${Task}$.
For each sampled task in $Task_{pre}$, we sample two disjoint sets: a \textit{Support Set} $\mathcal{S}$ and a \textit{Query Set} $\mathcal{Q}$, used for prototype computation and parameter optimization, respectively.

Given the RDB-induced graph $\mathcal{G}$ and the hybrid node attributes $\mathbf{X}^H$, we first sample a temporal ego-subgraph $\mathcal{G}_v$ centered at each target node $v$. 
The graph encoder $f_\theta$ then computes the representation of $v$ as:
\begin{equation}
\label{equ:hv}
\mathbf{h}_v = f_\theta(\mathcal{G}_v, v; \mathbf{X}^H),
\end{equation}
Subsequently, for each pseudo-class $c$ in the current task, we compute the class prototype $\boldsymbol{\mu}_c$ by averaging the embeddings of the support nodes belonging to class $c$:
\begin{equation}
\boldsymbol{\mu}_c= \frac{1}{|\mathcal{S}_c|} \sum_{v \in \mathcal{S}_c} \mathbf{h}_v,
\end{equation}
where $\mathcal{S}_c \subset \mathcal{S}$ denotes the subset of support nodes labeled with pseudo-class $c$.

Finally, to optimize the encoder, we calculate the meta-learning loss $\mathcal{L}$ on the \textit{Query Set} $\mathcal{Q}$ using the negative log-likelihood based on the Euclidean distance to the prototypes:
\begin{equation}
    \label{eq:meta_loss}
    \mathcal{L} = - \frac{1}{|\mathcal{Q}|} \sum_{(v, \tilde{y}) \in \mathcal{Q}} 
    \log \frac{
    \exp\left(- \| \mathbf{h}_v - \boldsymbol{\mu}_{\tilde{y}} \|_2^2 \right)
    }{
    \sum_{c} \exp\left(- \| \mathbf{h}_v - \boldsymbol{\mu}_{c} \|_2^2 \right)
    },
\end{equation}

Upon completion of pre-training, the optimized encoder $f_\theta$ is used to obtain the final graph-based representation for each target node $v \in \mathcal{V}_a$ according to Eq.~(\ref{equ:hv}).

\paragraph{\textbf{Adaptation.}}
For downstream adaptation, we adopt strategies according to both the supervision regime and the task type.
In the data-limited regime, for classification, we use non-parametric prototypical inference~\cite{liu2023pre}, which is aligned with the pre-training paradigm. Specifically, labeled examples are treated as the support set to compute class prototypes, and test nodes are classified based on their distances to these prototypes~\cite{chen2019closer}.
For regression, since prototype-based inference is not directly applicable to continuous targets, we fine-tune a lightweight regression head, such as a Multi-Layer Perceptron (MLP), on top of the pre-trained encoder~\cite{wang2025griffin}.
In the data-sufficient regime, we fine-tune a lightweight task-specific head for both classification and regression, allowing the model to better exploit the available supervision.

\section{Experiments and Analysis}
\label{sec:experiments and analysis}
To evaluate the effectiveness of RelPrism, we perform extensive experiments on 14 tasks across 5 real-world datasets to answer the following research questions:

\textbf{RQ1:} How does RelPrism perform against state-of-the-art baselines in data-limited scenarios?

\textbf{RQ2:} How does RelPrism perform against state-of-the-art baselines in data-sufficient scenarios?

\textbf{RQ3:} How does the quality of representations learned by RelPrism compare with those generated by other SSL methods?

\textbf{RQ4:} How do the clustering-generated pseudo-tasks perform relative to randomized pseudo-tasks (lower bound) and tasks derived from ground-truth labels (upper bound)?

\textbf{RQ5:} How do key components of RelPrism affect performance, including graph construction and the contributions of intrinsic, relational, and hybrid task pools?


Furthermore, we present a hyperparameter study in Appendix C~\footnote{The appendices are available in the code repository.}, a case study with detailed data samples in Appendix D, and analyses of time complexity and scalability in Appendix E.

\begin{table*}[tbp]
\centering
\caption{\textbf{Performance Comparison on \textit{50}-Shot Classification (ROC-AUC $\uparrow$).} Best results are bolded and second-best are underlined.}
\label{tab:cls_50shot}
\resizebox{\textwidth}{!}{%
\begin{tabular}{c|c|ccc|ccc|cc|cc}
\toprule
\textbf{Dataset} & \textbf{Task}
& \textbf{LightGBM} & \textbf{RelGNN} & \textbf{RelGT}
& \textbf{STUNT} & \textbf{GraphCL} & \textbf{GraphMAE}
& \textbf{Griffin} & \textbf{RT}
& \textbf{TVE} & \textbf{RelPrism} \\
\midrule

\multirow{2}{*}{\texttt{rel-f1}}
& \texttt{driver-top3}
& $0.729_{\pm 0.06}$ & $0.592_{\pm 0.15}$ & $0.591_{\pm 0.03}$
& $0.616_{\pm 0.01}$ & $0.585_{\pm 0.11}$ & $0.555_{\pm 0.12}$
& $0.618_{\pm 0.05}$ & $\underline{0.760_{\pm 0.00}}$
& $0.745_{\pm 0.02}$ & $\mathbf{0.765_{\pm 0.03}}$ \\

& \texttt{driver-dnf}
& $0.604_{\pm 0.05}$ & $0.536_{\pm 0.09}$ & $0.550_{\pm 0.04}$
& $0.519_{\pm 0.06}$ & $0.566_{\pm 0.08}$ & $0.512_{\pm 0.11}$
& $\underline{0.633_{\pm 0.03}}$ & $0.629_{\pm 0.02}$
& $0.569_{\pm 0.07}$ & $\mathbf{0.649_{\pm 0.07}}$ \\
\midrule

\multirow{2}{*}{\texttt{rel-stack}}
& \texttt{user-engagement}
& $0.545_{\pm 0.02}$ & $0.563_{\pm 0.04}$ & $0.586_{\pm 0.08}$
& $0.569_{\pm 0.05}$ & $0.522_{\pm 0.19}$ & $0.510_{\pm 0.05}$
& $0.637_{\pm 0.01}$ & $0.672_{\pm 0.02}$
& $\mathbf{0.739_{\pm 0.01}}$ & $\underline{0.689_{\pm 0.05}}$ \\

& \texttt{user-badge}
& $0.545_{\pm 0.01}$ & $0.559_{\pm 0.03}$ & $0.577_{\pm 0.04}$
& $0.564_{\pm 0.00}$ & $0.523_{\pm 0.17}$ & $0.496_{\pm 0.25}$
& $\underline{0.784_{\pm 0.02}}$ & $0.771_{\pm 0.02}$
& $0.756_{\pm 0.00}$ & $\mathbf{0.816_{\pm 0.02}}$ \\
\midrule

\multirow{2}{*}{\texttt{rel-amazon}}
& \texttt{user-churn}
& $0.501_{\pm 0.01}$ & $0.538_{\pm 0.01}$ & $0.565_{\pm 0.04}$
& $0.525_{\pm 0.01}$ & $0.493_{\pm 0.05}$ & $0.488_{\pm 0.02}$
& $0.494_{\pm 0.01}$ & $0.562_{\pm 0.06}$
& $\underline{0.624_{\pm 0.00}}$ & $\mathbf{0.631_{\pm 0.00}}$ \\

& \texttt{item-churn}
& $0.507_{\pm 0.01}$ & $0.521_{\pm 0.01}$ & $0.533_{\pm 0.03}$
& $0.526_{\pm 0.01}$ & $0.520_{\pm 0.05}$ & $0.506_{\pm 0.02}$
& $0.513_{\pm 0.02}$ & $\underline{0.703_{\pm 0.00}}$
& $0.635_{\pm 0.00}$ & $\mathbf{0.731_{\pm 0.00}}$ \\
\midrule

\texttt{rel-hm}
& \texttt{user-churn}
& $0.514_{\pm 0.01}$ & $0.556_{\pm 0.02}$ & $0.554_{\pm 0.01}$
& $0.519_{\pm 0.01}$ & $0.482_{\pm 0.04}$ & $0.524_{\pm 0.03}$
& $0.582_{\pm 0.04}$ & $0.583_{\pm 0.00}$
& $\underline{0.648_{\pm 0.14}}$ & $\mathbf{0.671_{\pm 0.00}}$ \\
\bottomrule
\end{tabular}%
}
\end{table*}

\begin{table*}[tbp]
\centering
\caption{\textbf{Performance Comparison on \textit{50}-Shot Regression (MAE $\downarrow$).} Best results are bolded and second-best are underlined.}
\label{tab:reg_50shot}
\resizebox{\textwidth}{!}{%
\begin{tabular}{c|c|ccc|ccc|cc|cc}
\toprule
\textbf{Dataset} & \textbf{Task}
& \textbf{LightGBM} & \textbf{RelGNN} & \textbf{RelGT}
& \textbf{STUNT} & \textbf{GraphCL} & \textbf{GraphMAE}
& \textbf{Griffin} & \textbf{RT}
& \textbf{TVE} & \textbf{RelPrism} \\
\midrule
\multirow{2}{*}{\texttt{rel-amazon}}
& \texttt{user-ltv}
& $2.854_{\pm 0.09}$ & $3.457_{\pm 0.11}$ & $3.181_{\pm 0.11}$
& $0.397_{\pm 0.07}$ & $0.338_{\pm 0.01}$ & $0.352_{\pm 0.01}$
& $0.370_{\pm 0.02}$ & $0.355_{\pm 0.05}$
& $\underline{0.313_{\pm 0.00}}$ & $\mathbf{0.302_{\pm 0.02}}$ \\

& \texttt{item-ltv}
& $0.780_{\pm 0.02}$ & $0.959_{\pm 0.08}$ & $0.797_{\pm 0.04}$
& $0.151_{\pm 0.03}$ & $0.167_{\pm 0.01}$ & $0.210_{\pm 0.02}$
& $0.182_{\pm 0.06}$ & $0.181_{\pm 0.07}$
& $\underline{0.144_{\pm 0.02}}$ & $\mathbf{0.132_{\pm 0.02}}$ \\
\midrule

\texttt{rel-f1}
& \texttt{driver-position}
& $0.755_{\pm 0.03}$ & $0.887_{\pm 0.18}$ & $0.795_{\pm 0.02}$
& $0.782_{\pm 0.02}$ & $\mathbf{0.677_{\pm 0.07}}$ & $\underline{0.681_{\pm 0.05}}$
& $0.689_{\pm 0.06}$ & $0.734_{\pm 0.10}$
& $0.762_{\pm 0.04}$ & $0.758_{\pm 0.04}$ \\
\midrule

\texttt{rel-hm}
& \texttt{item-sales}
& $2.223_{\pm 0.23}$ & $1.968_{\pm 0.08}$ & $1.772_{\pm 0.04}$
& $0.254_{\pm 0.02}$ & $0.230_{\pm 0.02}$ & $0.239_{\pm 0.02}$
& $2.365_{\pm 0.25}$ & $1.296_{\pm 0.13}$
& $\underline{0.189_{\pm 0.01}}$ & $\mathbf{0.163_{\pm 0.03}}$ \\
\midrule

\texttt{rel-stack}
& \texttt{post-votes}
& $1.137_{\pm 0.19}$ & $1.037_{\pm 0.06}$ & $1.289_{\pm 0.08}$
& $0.161_{\pm 0.02}$ & $\underline{0.134_{\pm 0.04}}$ & $0.186_{\pm 0.05}$
& $0.145_{\pm 0.04}$ & $0.137_{\pm 0.03}$
& $0.148_{\pm 0.03}$ & $\mathbf{0.124_{\pm 0.04}}$ \\
\midrule

\multirow{2}{*}{\texttt{rel-trial}}
& \texttt{study-adverse}
& $2.464_{\pm 0.14}$ & $2.766_{\pm 0.03}$ & $2.058_{\pm 0.14}$
& $0.279_{\pm 0.00}$ & $0.229_{\pm 0.02}$ & $0.252_{\pm 0.02}$
& $2.918_{\pm 0.10}$ & $2.367_{\pm 0.04}$
& $\underline{0.212_{\pm 0.02}}$ & $\mathbf{0.199_{\pm 0.02}}$ \\

& \texttt{site-success}
& $5.055_{\pm 0.08}$ & $4.797_{\pm 0.13}$ & $4.329_{\pm 0.11}$
& $0.980_{\pm 0.00}$ & $0.992_{\pm 0.11}$ & $0.970_{\pm 0.07}$
& $0.969_{\pm 0.01}$ & $\underline{0.940_{\pm 0.04}}$
& $1.031_{\pm 0.11}$ & $\mathbf{0.932_{\pm 0.04}}$ \\
\bottomrule
\end{tabular}%
}
\end{table*}

\subsection{Experimental Settings}

\textit{\textbf{Datasets.}}
We evaluate RelPrism on 14 tasks from RelBench~\cite{robinson2024relbench}, comprising 7 binary classification tasks and 7 regression tasks across 5 real-world datasets. These datasets span 4 domains: sports (\texttt{rel-f1}), social networks (\texttt{rel-stack}), medical records (\texttt{rel-trial}), and e-commerce (\texttt{rel-amazon} and \texttt{rel-hm}). 
They exhibit significant diversity in scale, with the number of tables ranging from 3 to 15, row counts spanning from tens of thousands to tens of millions, and column counts varying between 15 and 140.


\paragraph{\textbf{Baselines.}}
We compare RelPrism against nine representative baseline methods drawn from three distinct categories.

\begin{itemize}
    \item \textbf{Supervised Learning Methods:} We include LightGBM~\cite{ke2017lightgbm}, a widely used Gradient Boosting Decision Tree (GBDT) model for tabular data, and two end-to-end supervised RDL methods: RelGNN~\cite{chen2025relgnn} and RelGT~\cite{dwivedi2025relational2}.
    
    \item \textbf{Self-Supervised Learning Methods:} We compare RelPrism with STUNT~\cite{nam2023stunt}, a tabular SSL method; two GNN-based SSL methods, including the contrastive GraphCL~\cite{you2020graph} and the generative GraphMAE~\cite{hou2022graphmae}; and TVE~\cite{truong2025pre}, a recent SSL method specifically designed for RDBs. Following TVE and \citeauthor{liu2023flaky}, we adapt GraphCL and GraphMAE for RDB pre-training by incorporating tabular feature perturbations inspired by tabular SSL methods~\cite{yoon2020vime,nam2023stunt}.
    
    \item \textbf{Foundation Models:}
    To ensure a comprehensive evaluation, we include two state-of-the-art RDB foundation models, Griffin~\cite{wang2025griffin} and RT~\cite{ranjan2025relational}, as competitive baselines.
    For Griffin, we use the \textit{Griffin-RDB-SFT} checkpoint pre-trained on single-table and RDB datasets, and follow the authors' guidelines to select checkpoints whose pre-training data do not overlap with the target task, thereby preventing data leakage.
    For RT, we retrain the Transformer on the target dataset while excluding the target task to avoid leakage. More details are provided in Appendix A.
\end{itemize}

\paragraph{\textbf{Evaluation Protocols.}}
We consider two representative node-level predictive tasks in RDBs: classification and regression.
For data-limited scenarios, we evaluate few-shot settings with $k \in \{1, 5, 50\}$.
Following~\cite{robinson2024relbench,chen2025relgnn}, we report ROC-AUC for classification and MAE for regression.
For regression tasks, following Griffin's metric adjustment, MAE is computed on normalized target values.
To ensure consistency across baselines, all methods are evaluated on the normalized scale: most baselines are trained with pre-normalized targets, while LightGBM, RelGNN, RelGT, and RT perform normalization within their experimental pipelines.
For all experiments, we report the mean and standard deviation over five independent runs with different random seeds.

\paragraph{\textbf{Implementation Details.}}
\label{main:imple}
Following~\cite{wang2025griffin}, we use the pre-trained \textit{nomic-embed-text-v1.5} model to encode textual and categorical inputs, and use a pre-trained Multi-Layer Perceptron (MLP) to encode normalized numerical and temporal inputs.
For pseudo-task generation, the number of clusters $C_p$ is uniformly sampled from the range $[2, 20]$.
We adopt Graph Transformer~\cite{shi2020masked} as the graph encoder, which can naturally incorporate edge attributes.
For downstream classification in data-limited scenarios, each method follows the standard adaptation protocol corresponding to its training paradigm. Specifically, RelPrism performs non-parametric prototypical inference, while the baselines fine-tune task-specific heads with binary cross-entropy loss.
In data-sufficient scenarios, both RelPrism and the baselines fine-tune task-specific heads using binary cross-entropy loss.
For regression, all methods fine-tune a regression head with L1 loss in both scenarios.
Parameter initialization for all baselines follows the recommended settings in their original papers, with necessary hyperparameter search and optimization conducted for RDBs.
For GraphCL and GraphMAE, we adopt the tabular feature masking strategy from VIME~\cite{yoon2020vime} and uniformly set the feature masking rate to 0.3.
More implementation details are provided in Appendix B. 

\subsection{Data-Limited Performance (RQ1)}

\begin{figure*}[t]
    \centering
    \includegraphics[width=1\linewidth]{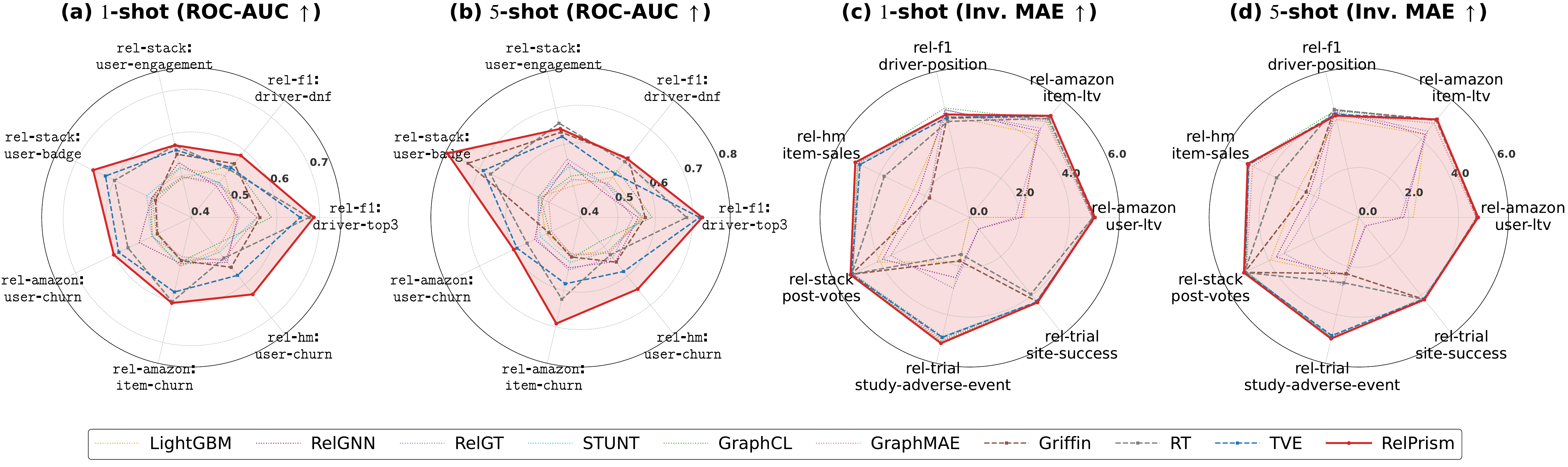}
    \caption{
    \textbf{1-Shot and 5-Shot Classification and Regression Performance on 14 Tasks across 5 Datasets.}
    For regression tasks, inverted MAE is used for visualization, where larger values indicate better performance.
   }
    \label{fig:fs_radar}
\end{figure*}

To evaluate model adaptability under limited supervision, we conduct experiments on 14 tasks across 5 datasets, including 7 classification tasks and 7 regression tasks. 
We report the \textit{1}-shot and \textit{5}-shot results for classification and regression in Figure~\ref{fig:fs_radar}, while the \textit{50}-shot results are presented in Tables~\ref{tab:cls_50shot} and~\ref{tab:reg_50shot}.
For radar-chart visualization, regression MAE values are inverted by subtracting them from the maximum MAE within the same shot setting, thereby aligning the visualization semantics with classification.
From these results, we draw the following key observations:

(1) Overall, RelPrism achieves state-of-the-art performance on 13 out of 14 tasks in data-limited scenarios across 5 datasets, outperforming the strongest baselines by an average of 4.15\% in ROC-AUC for classification and 10.75\% in MAE for regression.
(2) Supervised methods (LightGBM, RelGNN, and RelGT) perform poorly in label-scarce scenarios due to their reliance on extensive supervision.
(3) The tabular SSL and graph-based SSL baselines show clear limitations on RDBs. STUNT focuses on row-level intrinsic attributes and misses cross-table dependencies, while GraphCL and GraphMAE often underperform on classification because their graph SSL objectives may be misaligned with relational prediction~\cite{liu2023flaky}. However, their contrastive and reconstruction objectives preserve fine-grained node-level fidelity, making them competitive on regression tasks that favor detailed node-level information, e.g., \texttt{driver-position}.
(4) Griffin, RT, and TVE achieve competitive results, confirming the value of RDB pre-training. However, TVE focuses on subgraph-level relational structures, while Griffin and RT emphasize cell- or token-level fidelity through reconstruction. Their objectives thus remain tied to specific perspectives or granularities, leading to suboptimal performance.
(5) While mixed task sampling improves overall adaptability, it may slightly interfere with single-facet-dominated tasks, such as \texttt{user-engagement} and \texttt{driver-position}, where specialized baselines benefit from structural alignment or fine-grained feature fidelity. 
Nevertheless, RelPrism's balanced coverage across perspectives and granularities provides a more reliable foundation for downstream adaptation.

\subsection{Data-Sufficient Performance (RQ2)}
We further evaluate RelPrism in data-sufficient scenarios, following the original RelBench setting.
Specifically, we compare RelPrism with two end-to-end supervised RDL methods, RelGNN and RelGT, on four classification and three regression tasks, with results reported in Table~\ref{tab:cls_supervised}. 
We observe that supervised methods benefit substantially from abundant training examples and achieve performance comparable to RelPrism on classification tasks. 
However, RelPrism consistently performs better on regression tasks, suggesting that fine-grained continuous prediction benefits more from representations that capture comprehensive information across multiple perspectives and granularities.

\begin{table}[t]
\centering
\caption{\textbf{Performance Comparison in Data-Sufficient Scenarios.} Best results are bolded and second-best are underlined.}
\label{tab:cls_supervised}
\resizebox{\columnwidth}{!}{%
\begin{tabular}{c|c|ccc}
\toprule
\textbf{Dataset} & \textbf{Task} & \textbf{RelGNN} & \textbf{RelGT} & \textbf{RelPrism} \\
\midrule
\multicolumn{5}{c}{\textit{Classification (ROC-AUC $\uparrow$)}} \\
\midrule

\multirow{2}{*}{\texttt{rel-f1}}
& \texttt{driver-top3}
& $\underline{0.848_{\pm 0.01}}$
& $0.838_{\pm 0.03}$
& $\mathbf{0.850_{\pm 0.01}}$ \\

& \texttt{driver-dnf}
& $0.746_{\pm 0.02}$
& $\underline{0.749_{\pm 0.10}}$
& $\mathbf{0.756_{\pm 0.02}}$ \\
\midrule

\multirow{2}{*}{\texttt{rel-stack}}
& \texttt{user-engagement}
& $\mathbf{0.907_{\pm 0.00}}$
& $\underline{0.905_{\pm 0.00}}$
& $0.906_{\pm 0.00}$ \\

& \texttt{user-badge}
& $\mathbf{0.890_{\pm 0.00}}$
& $0.861_{\pm 0.00}$
& $\underline{0.884_{\pm 0.00}}$ \\
\midrule

\multicolumn{5}{c}{\textit{Regression (MAE $\downarrow$)}} \\
\midrule

\texttt{rel-f1}
& \texttt{driver-position}
& $\underline{0.638_{\pm 0.02}}$
& $0.655_{\pm 0.04}$
& $\mathbf{0.627_{\pm 0.02}}$ \\
\midrule

\texttt{rel-stack}
& \texttt{post-votes}
& $0.356_{\pm 0.00}$
& $\underline{0.360_{\pm 0.03}}$
& $\mathbf{0.120_{\pm 0.04}}$ \\
\midrule

\texttt{rel-trial}
& \texttt{study-adverse}
& $2.280_{\pm 0.02}$
& $\underline{1.799_{\pm 0.03}}$
& $\mathbf{0.152_{\pm 0.01}}$ \\
\bottomrule
\end{tabular}%
}
\end{table}

\subsection{Representation Quality Analysis (RQ3)}
\begin{figure}[t]
    \centering
    \includegraphics[width=1\linewidth]{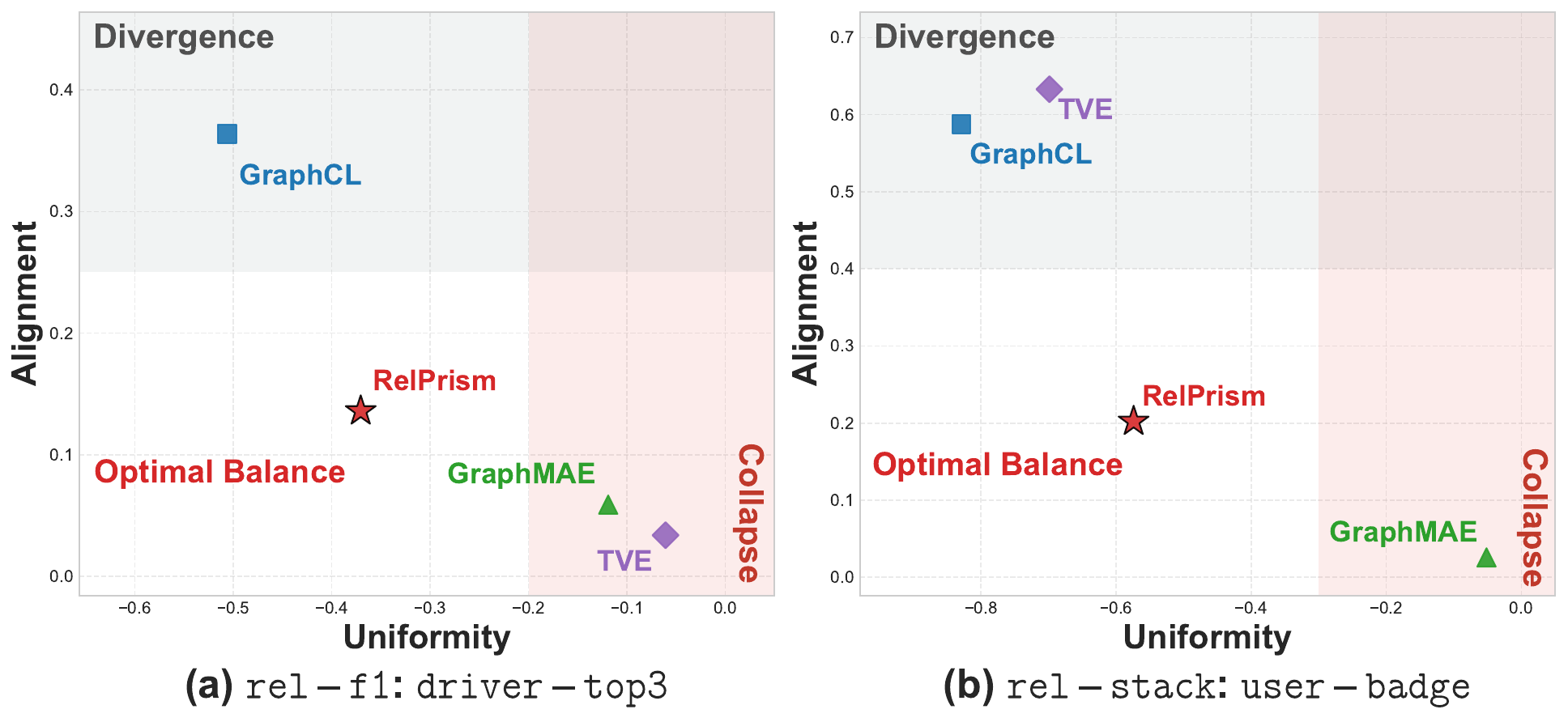}
    \caption{\textbf{Representation Quality Analysis via Alignment and Uniformity.} The bottom-left corner corresponds to ideal representations that are simultaneously compact within classes and globally diverse.  Our method achieves an optimal balance, avoiding both representation divergence and collapse.}
    \label{fig:repre}
\end{figure}

To assess the quality of the learned representations, we employ Alignment and Uniformity \cite{wang2020understanding} as cardinal metrics. Alignment constrains the local proximity of positive pairs, while Uniformity measures global feature diversity to preclude collapse. Optimal representations must minimize both simultaneously. Figure~\ref{fig:repre} depicts the results on two classification tasks (\textit{50}-shot): \texttt{driver-top3} and \texttt{user-badge}. The main findings are as follows:

(1) Our method achieves the optimal balance between the two metrics, residing in the ideal region (bottom-left). This demonstrates that our self-generated diverse tasks simultaneously optimize Alignment and Uniformity, yielding representations characterized by both intra-class compactness and global diversity.
(2) GraphCL exhibits representation divergence (low Uniformity, high Alignment). While distinguishing augmented views ensures distributional spread, the induced flaky information disrupts same-class aggregation. Conversely, our clustering-derived pseudo-labels provide explicit semantic supervision, significantly enhancing Alignment.
(3) GraphMAE exhibits representation collapse (low Alignment, high Uniformity), as reconstructing intrinsic attributes confines samples to a narrow cone, degrading discriminability. By contrast, our diverse pseudo-tasks compel the capture of multi-faceted information, effectively averting collapse.
(4) TVE exhibits extreme instability. It suffers from representation collapse on \texttt{driver-top3}, yet displays misalignment on \texttt{user-badge}. This underscores that pre-training on single-faceted information lacks the generalization capability to adapt to diverse distributions.

\subsection{Pseudo-Task Quality Analysis (RQ4)}
\label{main:task quality}
To rigorously assess the quality and semantic validity of clustering-generated pseudo-tasks, we compare them against two baselines across four tasks under the \textit{50}-shot setting. \textit{Randomized Pseudo-tasks} serve as a performance lower bound, assigning arbitrary pseudo-labels to randomly sampled nodes without meaningful information guidance. In contrast, \textit{Ground-Truth Tasks} serve as a theoretical upper bound and are constructed directly from manually annotated ground-truth labels. For regression tasks, continuous ground-truth values are discretized into pseudo-classes~\cite{torgo1997regression} to ensure compatibility with the prototypical pre-training paradigm.

\begin{figure}[tbp]
    \centering
    \includegraphics[width=1\linewidth]{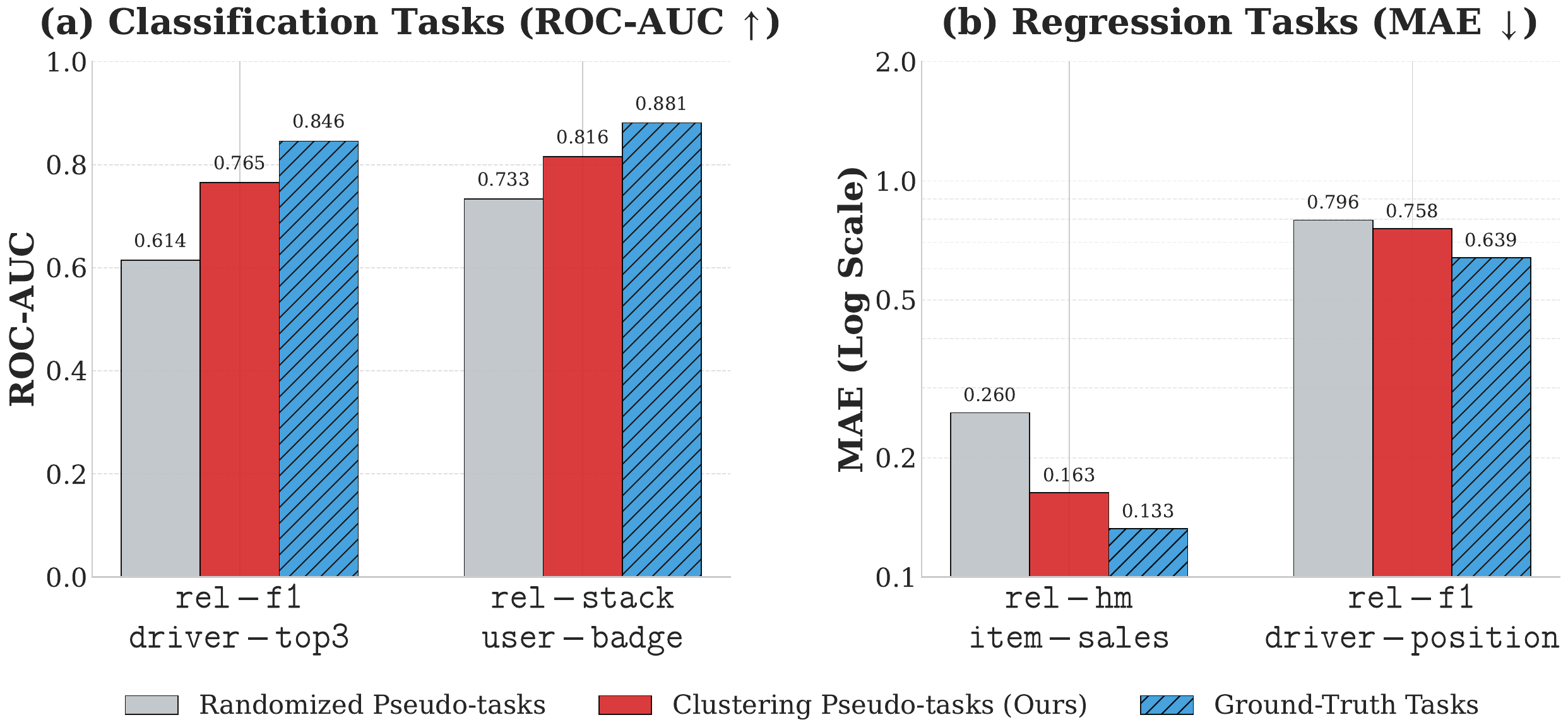}
    \caption{\textbf{Pseudo-Task Quality Analysis.} Our clustering-generated pseudo-tasks capture meaningful information comparable to ground-truth tasks that typically rely on expensive and labor-intensive manual labeling.}
    \label{fig:task}
\end{figure}

As shown in Figure~\ref{fig:task}, our method substantially outperforms the randomized baseline and narrows the gap to the ground-truth upper bound across both classification and regression tasks. Notably, despite lacking meaningful semantic supervision, the randomized baseline still achieves non-trivial performance with \textit{50} support samples, highlighting the adaptation capability fostered by multi-task training. Overall, these results demonstrate that our clustering-based task generation captures informative patterns approaching those provided by manual labels, consistent with prior work~\cite{hsu2018unsupervised}.

\begin{table}[t]
\centering
\caption{\textbf{Performance Comparison in Data-Sufficient Scenarios of Different RDB-to-Graph Constructions.} \textit{FN} denotes the RDL graph construction that treats fact-table rows as nodes, while \textit{FE} represents interactions inherent to fact tables as edges. Best results are bolded. Best results are bolded.}
\label{tab:graph_construction}
\resizebox{\columnwidth}{!}{%
\begin{tabular}{c|c|cc}
\toprule
\textbf{Dataset} & \textbf{Task} & \textbf{FN} & \textbf{FE (Ours)} \\
\midrule
\multicolumn{4}{c}{\textit{Classification (ROC-AUC $\uparrow$)}} \\
\midrule

\multirow{2}{*}{\texttt{rel-f1}}
& \texttt{driver-top3}
& $0.741_{\pm 0.03}$
& $\mathbf{0.850_{\pm 0.01}}$ \\

& \texttt{driver-dnf}
& $0.713_{\pm 0.01}$
& $\mathbf{0.756_{\pm 0.02}}$ \\
\midrule

\multirow{2}{*}{\texttt{rel-stack}}
& \texttt{user-engagement}
& $0.872_{\pm 0.03}$
& $\mathbf{0.906_{\pm 0.00}}$ \\

& \texttt{user-badge}
& $0.870_{\pm 0.00}$
& $\mathbf{0.884_{\pm 0.00}}$ \\
\midrule

\multicolumn{4}{c}{\textit{Regression (MAE $\downarrow$)}} \\
\midrule

\texttt{rel-f1}
& \texttt{driver-position}
& $0.659_{\pm 0.01}$
& $\mathbf{0.627_{\pm 0.02}}$ \\
\midrule

\texttt{rel-stack}
& \texttt{post-votes}
& $0.149_{\pm 0.01}$
& $\mathbf{0.120_{\pm 0.04}}$ \\
\midrule

\texttt{rel-trial}
& \texttt{study-adverse}
& $0.181_{\pm 0.01}$
& $\mathbf{0.152_{\pm 0.01}}$ \\

\bottomrule
\end{tabular}%
}
\end{table}

\subsection{Ablation Study (RQ5)}
\label{sec:task_pools}
We conduct ablation studies on graph construction and task pools. For graph construction, we compare our RDB-to-graph construction with RDL~\cite{fey2023relational}, which represents each fact-table row as a node, whereas ours models fact-table interactions as edges. This comparison covers four classification and three regression tasks in data-sufficient settings. For task pools, we evaluate the benefits of mixed multi-faceted pre-training by restricting meta-learning to the Intrinsic, Relational, or Hybrid Pool, respectively, under the \textit{1}-shot setting on two classification and two regression tasks.


As shown in Table~\ref{tab:graph_construction}, the fact-as-edge construction (ours) achieves better performance than the fact-as-node construction. By modeling fact-table rows as edges, it avoids introducing intermediate fact nodes, thereby reducing redundant aggregation and information mixing. It also better preserves the many-to-many interactions~\cite{chen2025relgnn} encoded in fact tables, leading to more effective relational message passing and improved performance.


The task-pool ablation results are shown in Table~\ref{tab:abla}. We observe a clear divergence in task preferences.
Classification tasks benefit most from the Hybrid Pool, which captures complementary intrinsic and relational information beyond any single perspective, while the Relational Pool also yields competitive results, highlighting the importance of relational interaction patterns.
In contrast, regression tasks favor the Intrinsic Pool, indicating their reliance on fine-grained node-specific information for continuous prediction. 
Overall, although single-pool pre-training can specialize to certain tasks, mixed sampling consistently ensures superior overall balance, validating the multi-faceted design of RelPrism. Furthermore, this observation could serve as an empirical guideline for calibrating the optimal task ratios during pre-training.

\begin{table}[t]
\centering
\caption{\textbf{Performance Comparison in \textit{1}-Shot under Different Task Pool Sampling Strategies.} Best results are bolded, second best are underlined.}
\label{tab:abla}
\setlength{\tabcolsep}{2pt}
\resizebox{\linewidth}{!}{%
\begin{tabular}{c|cccc}
\toprule
 & \textbf{Intrinsic} & \textbf{Relational} & \textbf{Hybrid} & \textbf{All (Ours)} \\
\midrule
\multicolumn{5}{c}{\textit{Classification (ROC-AUC $\uparrow$)}} \\
\midrule

\begin{tabular}{@{}c@{}}\texttt{rel-f1} \\ \texttt{driver-top3}\end{tabular} 
& $0.547_{\pm 0.08}$ 
& $0.661_{\pm 0.08}$ 
& \underline{$0.674_{\pm 0.03}$} 
& $\mathbf{0.686_{\pm 0.04}}$ \\
\midrule

\begin{tabular}{@{}c@{}}\texttt{rel-stack} \\ \texttt{user-badge}\end{tabular} 
& $0.509_{\pm 0.03}$ 
& $0.542_{\pm 0.04}$ 
& \underline{$0.608_{\pm 0.05}$} 
& $\mathbf{0.655_{\pm 0.04}}$ \\
\midrule

\multicolumn{5}{c}{\textit{Regression (MAE $\downarrow$)}} \\
\midrule

\begin{tabular}{@{}c@{}}\texttt{rel-hm} \\ \texttt{item-sales}\end{tabular} 
& $\mathbf{0.303_{\pm 0.05}}$ 
& $0.433_{\pm 0.05}$ 
& $0.511_{\pm 0.03}$ 
& \underline{$0.327_{\pm 0.04}$} \\
\midrule

\begin{tabular}{@{}c@{}}\texttt{rel-f1} \\ \texttt{driver-position}\end{tabular} 
& $\mathbf{0.927_{\pm 0.05}}$ 
& $1.255_{\pm 0.01}$ 
& $1.191_{\pm 0.01}$ 
& \underline{$1.180_{\pm 0.02}$} \\
\bottomrule
\end{tabular}%
}
\end{table}

\section{Conclusion}
In this paper, we identify a key mismatch between existing single-faceted pre-training objectives and the multi-faceted information required by RDB predictive tasks. 
To address this issue, we propose RelPrism, a multi-faceted self-supervised learning framework tailored for RDBs. 
By applying multi-granularity clustering to attributes constructed from different perspectives, RelPrism generates intrinsic, relational, and hybrid pseudo-task pools, providing a comprehensive informational foundation for pre-training. 
Extensive experiments on 14 tasks across 5 real-world datasets demonstrate that our task generation strategy better captures the diverse informational preferences of downstream RDB tasks, achieving effective performance under both data-limited and data-sufficient scenarios.

\begin{acks}
This work is supported in part by the Major Key Project of PCL (Grant No. PCL2025A09), the National Natural Science Foundation of China (No. 62550138, 62192784, 62572064, 62472329), and the Beijing Natural Science Foundation (No. 253004).
\end{acks}

\section{Ethical Considerations}
RelPrism is designed as a general-purpose representation learning framework for relational databases. We evaluate it on publicly available benchmark datasets without collecting new personal data or directly involving human participants. Nevertheless, relational data in practical applications may contain sensitive information and historical biases that could be inherited by learned representations, potentially affecting downstream predictions and decisions. For deployment, we recommend appropriate anonymization, fairness and privacy auditing, domain-specific validation, and human oversight.


\bibliographystyle{ACM-Reference-Format}
\bibliography{sample-base}

\end{document}